\documentclass[preprint,12pt]{elsarticle}

\usepackage{amssymb}

\usepackage{dblfloatfix}

\journal{Computerized Medical Imaging and Graphics}

\begin{document}

\begin{frontmatter}



\title{3D Convolutional Neural Networks for Tumor Segmentation using Long-range 2D Context}



\author[a]{Pawel~Mlynarski }
\author[a]{Herv\'e~Delingette}
\author[b]{Antonio~Criminisi}
\author[a]{Nicholas~Ayache}
\address[a]{Universit\'e C\^ote d'Azur, Inria Sophia Antipolis, France.}
\address[b]{Microsoft Research Cambridge, United Kingdom.}

\begin{abstract}
We present an efficient deep learning approach for the challenging task of tumor segmentation in multisequence MR images. In recent years, Convolutional Neural Networks (CNN) have achieved state-of-the-art performances in a large variety of recognition tasks in medical imaging. Because of the considerable computational cost of CNNs, large volumes such as MRI are typically processed by subvolumes, for instance slices (axial, coronal, sagittal) or small 3D patches. In this paper we introduce a CNN-based model which efficiently combines the advantages of the short-range 3D context and the long-range 2D context. To overcome the limitations of specific choices of neural network architectures, we also propose to merge outputs of several cascaded 2D-3D models by a voxelwise voting strategy. Furthermore, we propose a network architecture in which the different MR sequences are processed by separate subnetworks in order to be more robust to the problem of missing MR sequences. Finally, a simple and efficient algorithm for training large CNN models is introduced. We evaluate our method on the public benchmark of the BRATS 2017 challenge on the task of multiclass segmentation of malignant brain tumors. Our method achieves good performances and produces accurate segmentations with median Dice scores of 0.918 (whole tumor), 0.883 (tumor core) and 0.854 (enhancing core). Our approach can be naturally applied to various tasks involving segmentation of lesions or organs.
\end{abstract}


\end{frontmatter}



\section{Introduction}
\begin{figure}[!]
\centering
\includegraphics[width=1.0\textwidth]{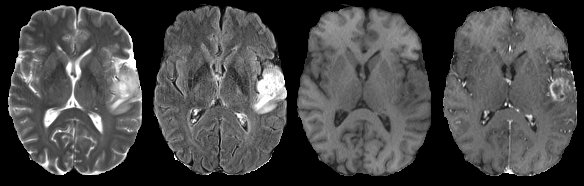}
\caption{Multisequence MR scan of a patient suffering from a glioblastoma. From left to right: T2-weighted, FLAIR, T1-weighted, post-contrast T1-weighted.}\label{fig_mri}
\end{figure} 

Gliomas are the most frequent primary brain tumors and represent approximatively 80\% of malignant brain tumors \cite{goodenberger2012genetics}. They originate from glial cells of the brain or the spine and can be classified according to the cell type, the grade and the location. High grade gliomas (grades III and IV) are associated with a particularly poor prognosis: patients diagnosed with glioblastoma multiforme survive on average 12-14 months under therapy. Medical images such as MRI \cite{bauer2013survey} are used for diagnosis, therapy planning and monitoring of gliomas.

Different tumor tissues (necrotic core, active rim, edema) can be imaged using multiple MR sequences. For instance, T2-FLAIR sequence is suitable for detecting edema while T1-weighted MR images acquired after the injection of a gadolinium-based contrast product are suitable to detect active parts of the tumor core (Fig. \ref{fig_mri}). These tumor tissues may be treated with different therapies \cite{gillies2015radiomics} and their analysis is important to assess the tumor characteristics, in particular its malignity.

Manual segmentation of tumors is a challenging and time-consuming task. Moreover, there is a significant variability between segmentations produced by human experts. An accurate automatic segmentation method could help in therapy planning and in monitoring of the tumor progression by providing the exact localization of tumor subregions and by precisely quantifying their volume.

Tumor variability in location, size and shape makes it difficult to use probabilistic priors. Image intensities of voxels representing tumor tissues in MR images highly overlap with intensities of other pathologies or healthy structures. Furthermore, ranges of MR image intensities highly vary from one imaging center to another depending on the acquisition system and the clinical protocol. Due to these aspects, in order to determine the presence of a tumor at a given position, high-level contextual information has to be analyzed.

A large variety of methods have been proposed for multiclass tumor segmentation. In 2012, the Multimodal Brain Tumor Segmentation Challenge (BRATS) \cite{menze2015multimodal,bakas2017advancing} was launched. The first group of methods corresponds to generative models based on the registration of the patient scan to a brain atlas providing a spatially varying probabilistic prior of different tissues. In the method of Prastawa et al \cite{prastawa2004brain}, tumor segmentation is guided by differences between the patient scan and the atlas of healthy brain. One limitation of this approach is the fact that it ignores the mass effect (deformation of neighboring healthy structures) caused by the tumor, which can lead to incorrect registration. In methods such as GLISTR \cite{gooya2012glistr} or \cite{kwon2014combining}, the authors propose to modify a healthy atlas by using tumor growth models and to perform a joint segmentation and registration to a modified brain atlas. These methods have the advantage of taking into account the characterics of tumors, however the use of tumor growth models comes with an additional complexity and the estimation of the number of tumor seeds is non trivial. A multi-atlas method, based on the search of similar image patches, was also proposed by Cordier et al \cite{cordier2016patch}.

Promising results were obtained by discriminative models corresponding to voxelwise classifiers such as SVM \cite{bauer2011fully,lee2005segmenting} or Random Forests \cite{ho1995random, zikic2012decision, geremia2012spatial, le2016lifted,bauer2012segmentation,tustison2015optimal}. For instance, Geremia et al  \cite{geremia2012spatial} propose to classify each voxel of a multimodal MR brain image by a random forest using features capturing information from neighbooring voxels and from distant regions such as the symmetric part of the brain. More recently, Le Folgoc et al proposed Lifted Auto-Context Forests \cite{le2016lifted}, an efficient method based on cascaded Random Forests progressively segmenting tumor subclasses exploiting the semantics of labels.

In recent years, Convolutional Neural Networks \cite{lecun1995convolutional} achieved state-of-the-art results in many tasks of image classification \cite{he2016deep, krizhevsky2012imagenet, simonyan2014very}, detection \cite{sermanet2013overfeat} and segmentation \cite{long2015fully,chen2014semantic}. In particular, the representation learning ability of CNNs is a considerable advantage for the task of tumor segmentation, where the design of discriminant image features is non trivial. The CNN-based methods of Pereira et al \cite{pereira2015deep} and Kamnitsas et al \cite{kamnitsas2016efficient} obtained respectively the best performance in BRATS 2015 and BRATS 2016 challenges. Fully-convolutional neural networks \cite{long2015fully,ronneberger2015u,havaei2017brain,zheng20183d} were used in most state-of-the-art segmentation methods, in particular, recently we observe a particular interest for 3D fully-convolutional neural networks \cite{dou20173d,cciccek20163d, kamnitsas2017ensembles, wang2017automatic,isensee2017brain}. Many methods include postprocessing steps, often based on Conditional Random Fields \cite{lafferty2001conditional} or mathematical morphology \cite{serra2012mathematical}.

Despite promising results obtained by these methods, segmentation of tumors in large medical images is still a very challenging task. One of the main drawbacks of CNNs is their computational cost resulting from application of thousands of costly operations (convolutions, poolings, upsamplings) on input images. This aspect is particularly problematic for segmentation problems in large medical images such as MRI or CT scans. Despite the variety of proposed neural network architectures, current CNN-based systems struggle to capture a large 3D context from input images. Moreover, most methods implicitly assume the presence of all MR sequences for all patients and the correct registration between sequences whereas these conditions do not necessarily hold in practice. 

In this paper we analyze the drawbacks of commonly used CNN-based models and we propose an efficient system based on a 2D-3D model in which features extracted by 2D CNNs (capturing a rich information from a long-range 2D context in three orthogonal directions) are used as an additional input to a 3D CNN.

First, we propose a 2D model (processing axial, coronal or sagittal slices of the input image) in which we introduce an alternative approach for treating different MR sequences. In many CNNs, including the state-of-the-art deep learning models mentioned before, all channels of the input MR image are directly combined by the first convolutional layers of the network. We propose an architecture composed of modality-specific subnetworks (which can be trained independently) and of a joint part combining all input modalities. Such design allows to train one part of the network on images with missing MR sequences while also extracting a rich information resulting from the combination of all MR sequences. Furthermore, models processing images by slices do not assume the fixed resolution in three dimensions accross patients and therefore can be trained on larger databases than 3D models.

We propose to use features learned by 2D CNNs as an additional input to a 3D CNN in order to capture rich information extracted from a very large spatial context while bypassing computational constraints. Such design considerably increases the size of the receptive field compared to standard 3D models taking as input only the raw intensities of voxels of a subvolume.

In order to combine the strengths of different network architectures, we introduce a voxelwise voting strategy to merge multiclass segmentations produced by several models. Finally, we designed a simple and stable training algorithm which is particularly well adapted for training large models. 

We have evaluated our method on the challenging task of multiclass tumor segmentation of malignant brain tumors in multisequence MR images from the validation set of BRATS 2017 challenge, using a public benchmark. In the performed experiments, our 2D-3D approach has outperformed the standard 3D model (where a CNN takes as input only the raw intensities of voxels of a subvolume) and our system has obtained promising results with median Dice scores of 0.918, 0.883 and 0.854 respectively for the three tumor subregions considered in the challenge (whole tumor, tumor core and contrast-enhancing core). Our method can be adapted to a large variety of multiclass segmentation tasks in medical imaging.

\section{Methods}
\begin{figure}[h!]
\centering
\includegraphics[width=0.89\textwidth]{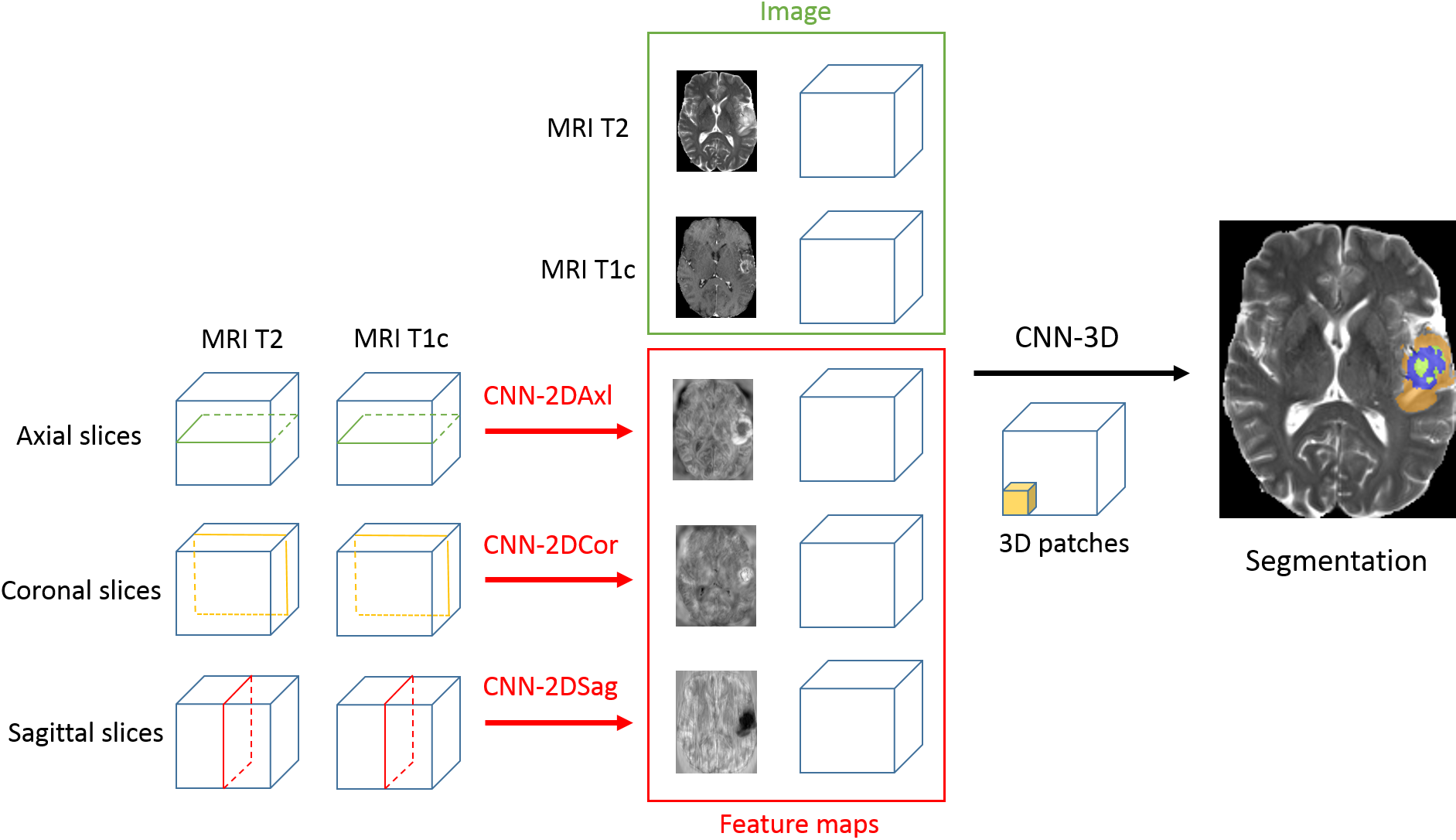}
\caption{Illustration of our 2D-3D model. Features extracted by 2D CNNs (processing the image by axial, coronal and sagittal slices) are used as additional channels of the patch processed by a 3D CNN. As these features encode a rich information extracted from a large spatial context, their use significantly increases the size of the receptive field of the 3D model.}
\label{fig_method}
\end{figure}

Our generic 2D-3D approach is illustrated on Fig. \ref{fig_method}.

The main components of our method are described in the following. First, we introduce an efficient 2D-3D model with a long-range 3D receptive field. Second, we present our neural network architecture with modality-specific subnetworks. In order to be more robust to limitations of specific choices of neural network architectures, we propose a voxelwise voting strategy to merge segmentations produced by several models.  Finally, we present a simple and efficient optimization algorithm.

\subsection{Spatial context and 3D models}
\label{model_3D}
A typical multisequence MR scan is composed of several millions of voxels. Convolutional neural networks transform input images by applying hundreds of convolutions and other operations whose outputs have to be stored in memory during iterations of the training in order to compute gradients of the loss by Backpropagation algorithm \cite{dreyfus1990artificial}. Training of CNNs requires typically dozens of thousands of iterations. Because of the cost of CNNs in time and memory, large medical images are generally processed by subvolumes of limited size. 

\begin{figure}[h!]
\centering
\includegraphics[width=0.82\textwidth]{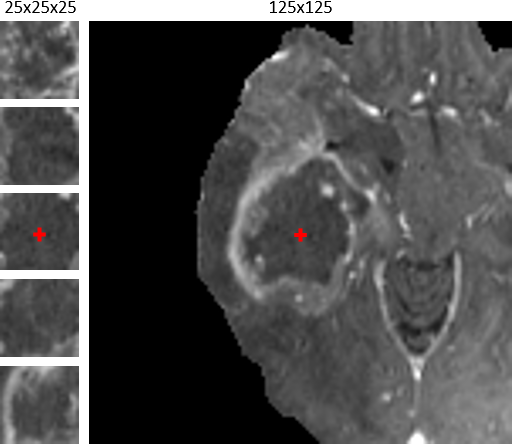}
\caption{Comparison of information represented by a 25x25x25 patch (left: 5 slices shown) and a 125x125 axial 2D patch centered at the same point. While both patches have the same number of voxels, the spatial context is considerably different. While the first patch captures local 3D shapes, the second patch captures information from distant points within the same plane.}
\label{fig_patch_slice}
\end{figure}

The obvious limitation of standard 2D approaches is to ignore one spatial dimension. However networks processing images by planes (axial, coronal or sagittal) have the ability to compare a studied voxel with distant voxels within the same plane and to capture a relevant information while keeping the input size reasonable. In the single-scale setting, the choice between the 2D and 3D option can therefore be seen as the choice between comparing distant voxels within the same plane (long-range 2D context) or comparing close voxels in three dimensions (short-range 3D context). Fig. \ref{fig_patch_slice} depicts the comparison of the information represented by a 2D patch of dimensions 125x125 and a 3D patch of dimensions 25x25x25 (both having the same number of voxels).

Another option is to process three orthogonal planes and classify the voxel at the intersection of three planes. This approach was successfully aplied by Ciompi et al. \cite{ciompi2017towards} for the problem of classification of lung nodules. The system proposed by the authors is composed of 9 streams processing 2D patches in three orthogonal planes centered at a givel voxel and at three different scales. The streams are then combined by fully-connected layers with the last layer performing classification. Unfortunately this system is computationally inefficient for the segmentation task because it implies redundant computations and reading operations on voxels which are in the same plane. As consequence, only few voxels of the ground truth could be considered at each iteration of the training whereas systems performing dense segmentation (classifying simultaneously several voxels) can be trained on several thousands of voxels in each iteration.

A larger 3D context can be analyzed by extracting multiscale 3D patches as in Deep Medic \cite{kamnitsas2016efficient}, a state-of-the-art CNN-based system which processes two-scale 3D patches by two streams of convolutional layers. The main characteristic of this design is the separate processing at two scales. A more global information is captured by the stream processing the patch from the image downsampled by a factor 3. However, this global information is not of the same nature as the one extracted by U-net \cite{ronneberger2015u} in which it results from a long sequence of convolutions and max-poolings starting from the original scale of the image (from local and low-level information to global and high-level information). A possible limitation of the model is its sequential aspect: the only concatenation is before the two last hidden layers of the network whereas skip-connections seem to improve the performance of neural networks \cite{he2016deep}.

The idea of our 2D-3D approach is to take into account a very large 3D context by using features learned by 2D networks rather than simply processing downsampled versions of the input image. In fact, features learned by 2D CNNs encode a rich information extracted from a large spatial context and the use of these features allows to considerably increase the size of the receptive field of the model.

In our method we use fully-convolutional neural networks \cite{long2015fully}. A network processes the input image by a sequence of spatially-invariant transformations in order to output voxelwise classification scores for all classes. The outputs of transformations at the same level of processing form a \textit{layer} which can be seen as a multi-channel image when arranged in a grid as in commonly used deep learning libraries such as Theano \cite{bergstra2010theano} or TensorFlow \cite{abadi2016tensorflow}. In 3D CNNs, each layer of the network corresponds to a multi-channel image with three spatial coordinates. A convolutional layer whose number of feature maps is equal to the number of classes and whose ouput is penalized during the training is called \textit{classification layer}. The channels of a layer are called \textit{feature maps} whose points represent \textit{neurons}. The set of voxels in the input layer which are taken into account in the computation of the output of a given neuron is called the \textit{receptive field} of the neuron. 

\begin{figure*}[b!]
\centering
\includegraphics[width=1.0\textwidth]{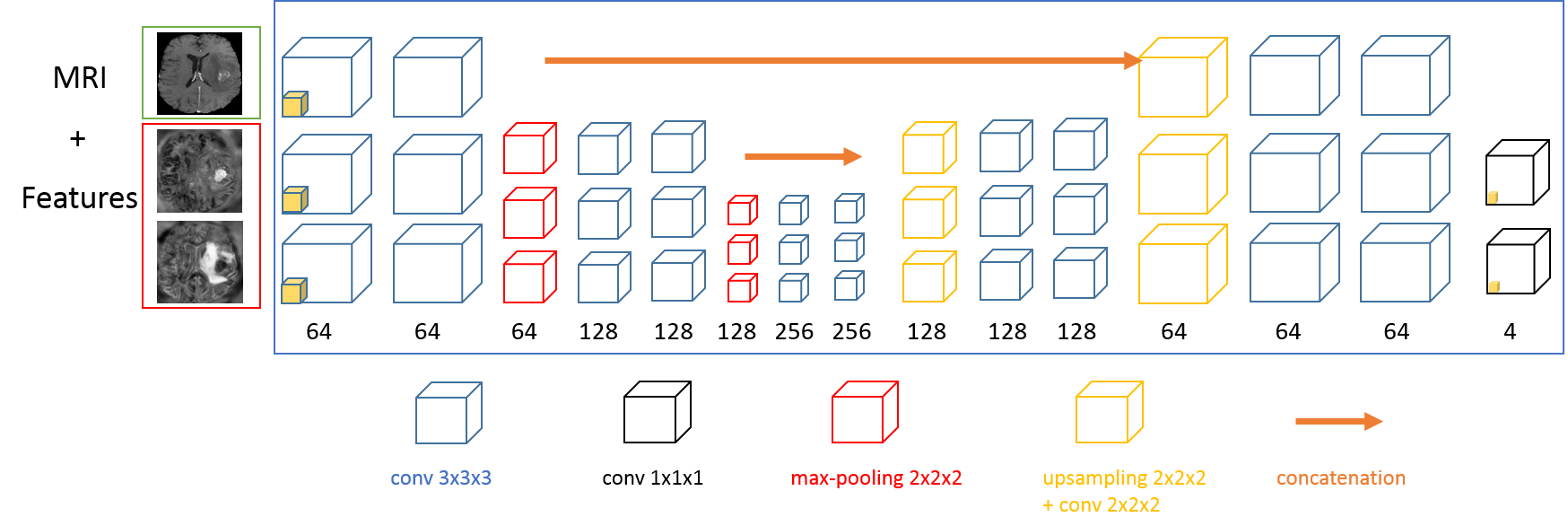}
\caption{Architecture of the main 2D-3D model used in our experiments (named '2D-3D model A' in the remainder). The number of feature maps in the last convolutional layer is equal to the number of classes (4 in our case).}
\label{fig_model3D}
\end{figure*}

\begin{figure}[h!]
\centering
\includegraphics[width=1.0\textwidth]{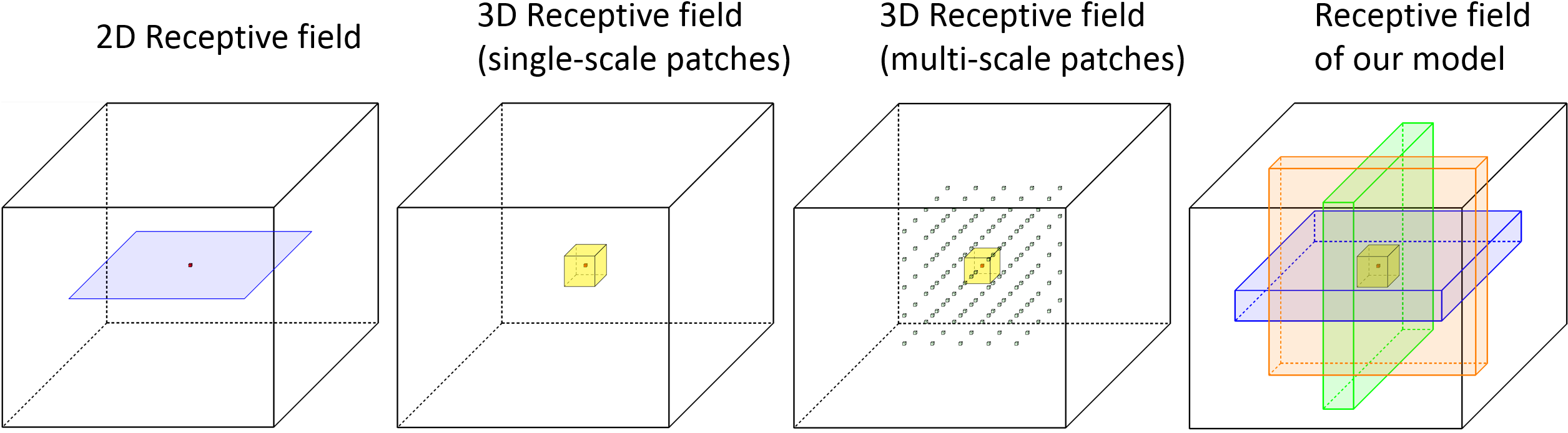}
\caption{Illustration of the receptive field of our 2D-3D model and the comparison with other approaches. The use of features extracted by 2D CNNs significantly increases the size of the receptive field compared to standard 3D approches which only use raw intensities of voxels of a subvolume.}
\label{fig_receptive_field}
\end{figure}

Our generic 2D-3D model (Fig.~\ref{fig_model3D}) is similar to 3D U-Net \cite{cciccek20163d} whose input is a 3D patch of the image along with a set of feature maps produced by networks trained on axial, coronal and sagittal slices (three versions of one 2D network). The extracted feature maps are concatenated to the input patch as additional channels. In our experiments we use three variants of this 2D-3D model, varying the way the extracted 2D features are imported in the network (section \ref{section_results}). The network processes 3D patches of size 70x70x70 and has the receptive field of size 45x45x45. However, given that the network takes as input not only the raw intensities of voxels but also the values of features extracted by 2D neural networks analyzing a large spatial context, the effective receptive field of the 2D-3D model is strikingly larger. Each feature represents a semantic information extracted from a large patch in axial, coronal or sagittal plane. The model uses the values of these features computed for all voxels. Therefore, classification of one voxel is performed using not only the raw intensities of voxels within the surrounding 45x45x45 patch but also from all axial, coronal and sagittal planes passing by the voxels of this patch (Fig. \ref{fig_receptive_field}). To the best of our knowledge, this is a novel way to capture a large 3D context with CNNs. The idea of using outputs of a CNN as additional input to another CNN was recently used for tumor segmentation in the work of Havaei et al \cite{havaei2017brain}, however the system proposed in \cite{havaei2017brain} is significantly different from our 2D-3D approach, in particular as it processes the image by axial slices, considered independently from each other.

The steps of the training of our model are the following:
\begin{enumerate}
\item Train three versions of the 2D network respectively on axial, coronal and sagittal slices. We refer to these three versions respectively as CNN-2DAxl, CNN-2DCor and CNN-2DSag, according to the nature of the captured 2D context.
\item For all images of the training database, extract the learned features from final convolutional layers (without softmax normalization) of the 2D neural networks (CNN-2DAxl, CNN-2DCor and CNN-2DSag) and save their outputs in files.
\item Train the 3D model using the extracted 2D features as additional channels to the input image patches.
\end{enumerate}

The choice of extracting features from the last convolutional layer is motivated by the fact that this layer has the largest receptive field and represents a semantic information while being composed of a small number of feature maps.

Given a training batch $b$ and the model parameters $\theta$, the loss function penalizes the output of the classification layer:
\begin{equation}
  Loss_{b}^{3D}(\theta)= -\frac{1}{V} \sum_{ i=1}^{|b|} \sum_{ (x,y,z)} w_{ (x,y,z)}^i \log(p^l_{i, (x,y,z)}(\theta))
\end{equation}

where $V$ is the total number of voxels in the ground truth of the training batch, $w_{ (x,y,z)}^i$ is the weight associated to the voxel at the position (x,y,z) in the $i^{th}$ image of the batch and $p^l_{i, (x,y, z)}(\theta)$ is the classification softmax score given by the network to the ground truth label $l$ of this voxel. The purpose of using weights is to counter the problem of the severe class imbalance, tumor subclasses being considerably under-represented. In contrast to common approaches, in our loss function the weights $w_{ (x,y,z)}^i$ of the voxels are set automatically depending on the composition of the batch (number of examples of each class greatly varies accross image patches). We suppose that in each training batch there is at least one voxel of each class. Let's note $C$ the number of classes and $N_b^c$ the number of voxels of the class $c$ in the batch $b$. For each class $c$ we set a target weight $t_c$ with $0 \leq t_c \leq 1$ and $ \sum_{ c=0}^{C-1} t_c= 1$. Then all voxels of the class $c$ are assigned the weight $t_c/N_b^c$ so that the total sum of their weights accounts for the proportion $t_c$ of the loss function. To better understand the effect of this parameter, note that in the standard non-weighted cross-entropy each voxel has a weight of 1 and the total weight of the class $c$ is proportional to the number of voxels labeled $c$. It implies that setting a target weight $t_c$ larger than the proportion of voxels labeled $c$ increases the total weight of the class $c$ (favoring its sensitivity) and conversely. In our experiments, a satisfactory performance was obtained with the following target weights (chosen empirically): ${t_0=0.4}$, $t_1=0.2$, $t_2=0.2$, $t_3=0.2$, corresponding respectively to 'non-tumor', 'non-enhancing core', 'edema' and 'enhancing core' classes. The choice of these values has an influence on the sensitivity to different tumor subclasses, however, the final segmentation performance in terms of Dice score was not found to be very sensitive to these hyperparameters. We fixed a higher target weight for the non-tumor class to limit the risk of oversegmentation. However, given that non-tumor voxels represent approximately 98-99\% of voxels of the batch, we significantly decreased the weight of the non-tumor class compared to a standard cross-entropy loss (0.98 vs 0.4).

We didn't observe the need of using batch normalization \cite{ioffe2015batch} in our model as we obtain a satisfactory convergence of the training with our algorithm described in section \ref{section_training}. A possibly important aspect could be the architecture of used neural networks: deep and sequential architectures (without concatenations between layers at differents depths) could need batch normalization or other techniques controlling the ranges of values of computed features. However, we have normalized the input images to approximatively match the ranges of values of extracted 2D features.

\subsection{2D deep learning model}
\label{model_2D}
\begin{figure*}[h!]
\centering
\includegraphics[width=1.0\textwidth]{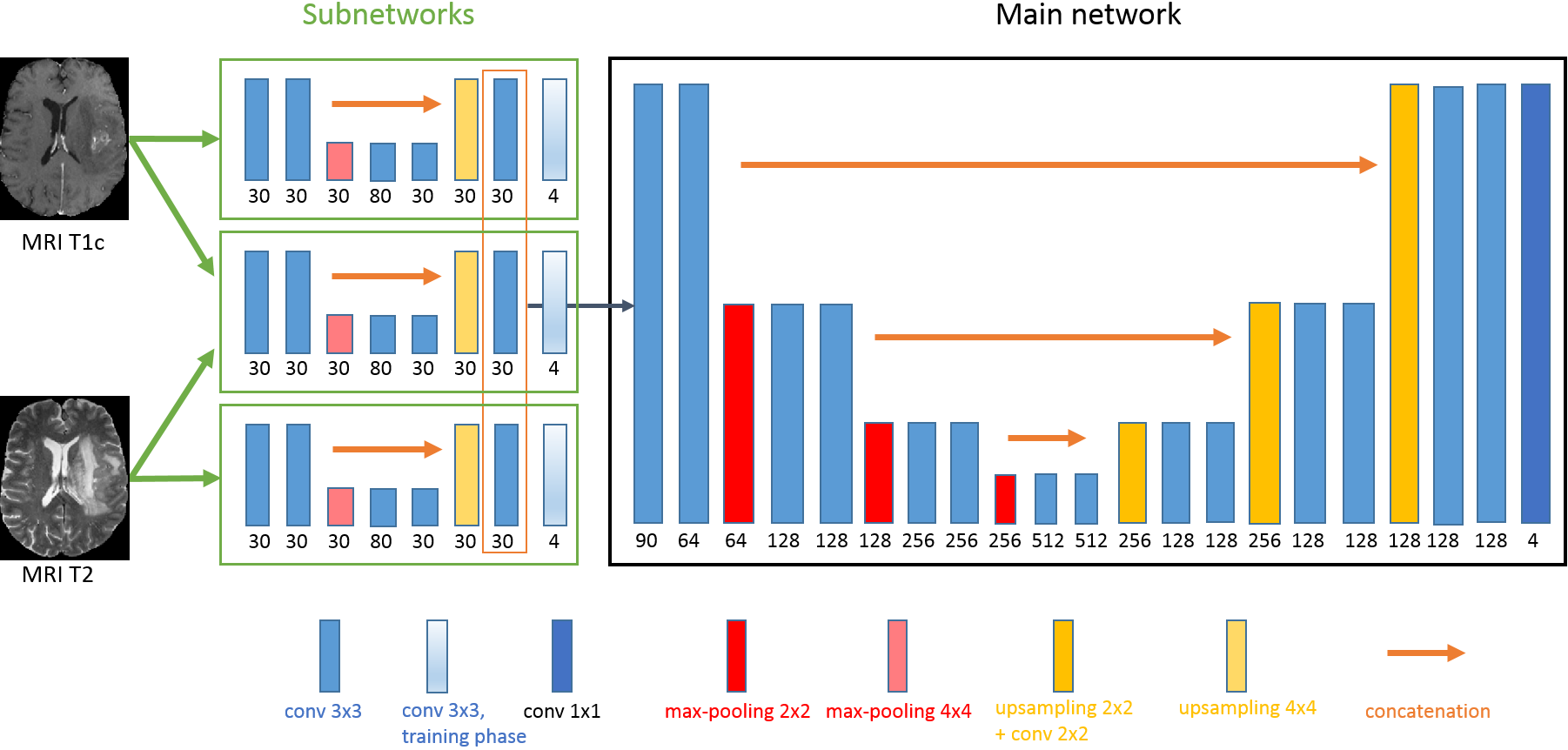}
\caption{Architecture of the main 2D model used in our experiments (named '2D model 1' in the remainder). The numbers of feature maps are specified below rectangles representing layers. In each subnetwork the first layer is concatenated to an upsampling layer in order to combine local and global information. Each subnetwork learns features specific to one image modality, except one subnetwork which directly combines all modalities. The classification layers of subnetworks are ignored during the test phase. For clarity purposes, we display the case with two MR sequences.}
\label{fig_model}
\end{figure*}

Our generic 2D deep learning model performs segmentation of tumors in axial, coronal or sagittal slices of a multisequence MRI. Our 2D model is similar to U-net \cite{ronneberger2015u} in which we introduce a system of co-trained subnetworks processing different input MR sequences (Fig. \ref{fig_model}). This design can be seen as a hybrid approach in which one part of the network processes independently different MR sequences and another part extracts features resulting from the combination of all sequences. Independent processing of input channels has the considerable advantage of being more robust to missing data. On the other hand, models using data from all input channels can extract important information resulting from relations between channels and therefore are likely to obtain better segmentation performance. Our goal is to combine these two aspects.

Given an input image with K channels, we consider K+1 subnetworks: one subnetwork per input channel and one subnetwork directly combining all channels. The subnetworks learn therefore features specific to each MR sequence (except the last subnetwork which learns features related to the direct combination of sequences) and can be trained on images with missing MR sequences. 

During the training phase we attach a classification layer to each subnetwork: more precisely, if a subnetwork has $n$ layers, then during the training phase we add one convolutional layer whose number of feature maps is equal to the number of classes and whose input is the $n^{th}$ layer of the subnetwork. The outputs of these additional layers, which we call \textit{auxiliary classification layers}, are penalized during the training according to eq. \ref{eq_loss2D} in order to force the subnetworks to extract the most pertinent information from each MR sequence. If the training database contains images with missing MR sequences, each modality-specific subnetwork can be pretrained independently of the others, on images for which the given MR sequence is provided. During the test phase, the auxiliary classification layers are ignored. The idea of using of intermediate losses to perform \textit{deep supervision} was succesfully used in the method of Dou et al \cite{dou20173d} for the problems of liver segmentation and vessel segmentation in 3D medical images.

In our experiments we considered two versions of subnetworks (section \ref{section_results}): in the first variant the subnetworks correspond to reduced versions of U-net and in the second variant they are composed of three convolutional layers (shallow version).

Final convolutional layers of the subnetworks are concatenated and fed to the main part of the network similar to U-net \cite{ronneberger2015u}. The main network is composed of two sections connected by concatenations of feature maps between layers at the same scale. The downsampling section is composed of convolutions and max-poolings. The upsampling section is composed of bilinear upsamplings, convolutions and concatenations with feature maps from the downsampling part. We apply batch normalization \cite{ioffe2015batch} in all convolutional layers except the classification layers. 

Given a training batch $b$ and the parameters $\theta$ of the network, the loss function penalizes the output of the final classification layer and the outputs of all auxiliary classification layers. 

For a classification layer L, the loss on the training batch $b$ corresponds to a pixelwise weighted cross-entropy given by:
\begin{equation}
  Loss_{b}^{L}(\theta)= -\frac{1}{P} \sum_{ i=1}^{|b|} \sum_{ (x,y)} w_{ (x,y)}^i \log(p^l_{i, (x,y)}(\theta))
\end{equation}
where $P$ is the total number of pixels in the ground truth of the training batch, $w_{ (x,y)}^i$ is the weight associated to the pixel at the position (x,y) in the $i^{th}$ image of the batch and $p^l_{i, (x,y)}(\theta)$ is the classification softmax score given by the network to the ground truth label $l$ for this pixel. The weights of pixels are adapted using the same strategy as in the 3D case. In the 2D case we used the following target weights: $t_0=0.7$, $t_1=0.1$, $t_2=0.1$, $t_3=0.1$.

The final loss is a convex combination of all intermediate losses, associated respectively with the main network and all subnetworks:

\begin{equation}
\label{eq_loss2D}
Loss_{b}(\theta)= c^{main} Loss_{b}^{main}(\theta)+ \sum_{k=1}^{K+1} c^{k} Loss_{b} ^{k}(\theta)
\end{equation}
where $K$ is the number of input channels, $0 \leq c^{main} \leq 1$, $0 \leq c^{k} \leq 1$ $\forall$ $k \in [1..K+1]$ and ${c^{main} +\sum_{k=1}^{K+1} c^{k}=1}$. In our experiments: $c^{main}=0.75$,  $K=4$ , $c^{k}=0.05$  ${\forall k \in [1..K+1]}$.

\subsection{Ensembles of neural networks}
\label{section_ensembling}
In order to be robust to limitations of particular choices of neural network architectures (kernels, strides, connectivity between layers, numbers of features maps, activation functions) we propose to combine segmentations produced by several models. The final segmentation is obtained by a voxelwise voting strategy exploiting the following relations between tumor subclasses: 
\begin{itemize}
\item Whole tumor region includes tumor-induced edema (class 2) and tumor core
\item Tumor core region includes contrast-enhancing core (class 3) and non-enhancing core (class 1)
\end{itemize}

\begin{figure}[!ht]
\centering
\includegraphics[width=0.75\textwidth]{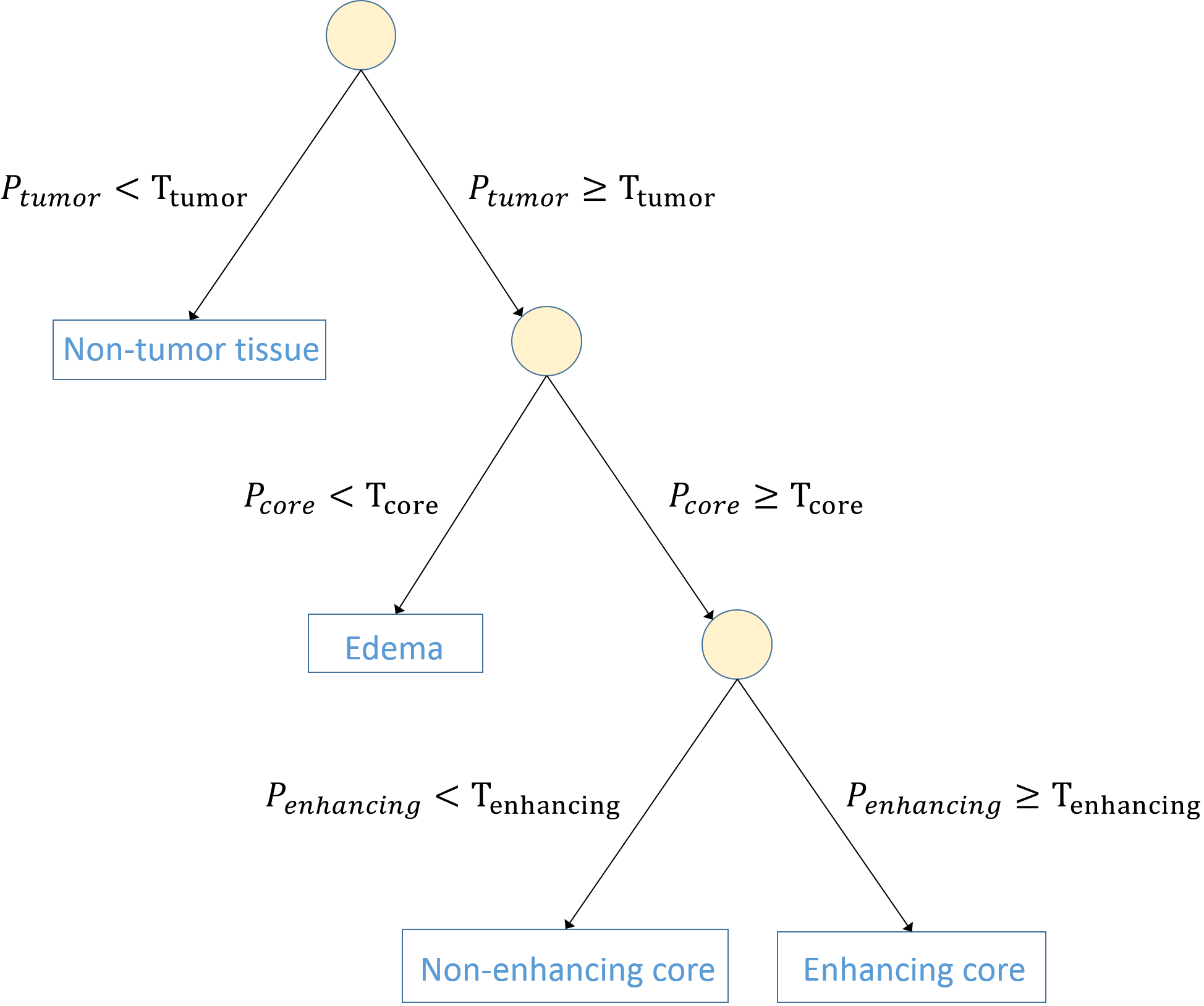}
\caption{Tree representing our decision process: leaves represent classes and nodes represent decisions according to aggregated votes for tumor subregions. The class of a voxel is progressively determined by thresholding on proportions of models which voted for given subregions.}
\label{fig_decision}
\end{figure}

Suppose we have $n$ tumor segmentations produced by different models (in our experiments n=6) and let's note $v_c$ the number of models which classified voxel $(x,y,z)$ as belonging to the class $c$, with $c \in \{0, 1, 2, 3\}$. The main idea is to aggregate the votes for classes according to their common regions and to take the decision in the hierarchical order, progressively determining the tumor subregions. We define the following quantities:
\begin{itemize}
\item $P_{tumor}=(v_1+v_2+v_3) / (v_0+ v_1+v_2+v_3)$ (proportion of votes for the whole tumor region in the total number of votes)
\item $P_{core}= (v_1+v_3)/(v_1+v_2+v_3)$ (proportion of votes for the 'tumor core' region among all votes for tumor subclasses)
\item $P_{enhancing}=v_3/ (v_1 + v_3)$ (proportion of votes for the contrast-enhancing core among all votes for the tumor core)
\end{itemize}
The decision process can be represented by a tree (Fig. \ref{fig_decision}) whose internal nodes represent the application of thresholding on the quantities defined above and whose leaves represent classes (final decision). The first decision is therefore to determine if a given voxel represents a tumor tissue, given the proportion of networks which voted for one of the tumor subclasses. If this proportion is above a chosen threshold (for example 0.4), we consider that the voxel represents a tumor tissue and we apply the same strategy to progressively determine the tumor subclass.

For each internal node $R$ (corresponding to a tumor subregion) of the decision tree, we therefore have to choose a threshold $T_R$ with $0<T_R \leq 1$. A high $T_R$ implies that a large proportion of models have to vote for this tumor subregion in order to consider its presence. The choice of this threshold therefore allows the user to control the trade-off between sensitivity and specificity of the corresponding tumor subregion. A low threshold gives priority to the sensitivity while a high threshold gives priority to the specificity. These thresholds can be chosen by cross-validation, observing the rate of non-detections and false positives. In our case, in order to increase the sensitivity to rare tumor subregions, we chose the following thresholds: $T_{tumor}=0.4$, $T_{core}=0.3$, $T_{enhancing}=0.4$.

\subsection{Training scheme}
\label{section_training}
Our training algorithm is a modified version of Stochastic Gradient Descent (SGD) with momentum \cite{rumelhart1988learning}. In each iteration of the standard SGD with momentum, the loss is computed on one batch $b$ of training examples and the vector $v$ of updates is computed as a linear combination of the previous update and the gradient of the current loss with respect to the parameters of the network: $ v^{t+1}=\mu v^{t} - \alpha_t \nabla  Loss_b (\theta^{t})  $ where $\theta^{t}$ are the current parameters of the network, $\mu$ is the momentum and $\alpha_t$ is the current learning rate. The parameters of the network are then updated:  $\theta^{t+1}= \theta^{t} + v^{t+1}$. We apply two main modifications to this scheme.

First, in each iteration we minimize the loss over several training batches instead of considering one batch per training iteration. Indeed, batches which can be processed by a large neural network (as the ones used in our approach) can be too small to correctly represent the training database, which will result in high oscillations of the loss and a difficult convergence. If we note $N$ the number of training batches per iteration, the loss at a given iteration is given by $Loss^{N}(\theta)~=~\sum_{b=1}^N Loss_b(\theta)$ where $Loss_b(\theta)$ is the loss over one training batch. Given the linearity of derivatives, the gradient of this loss with respect to the parameters of the network is simply the sum of gradients of losses over the N training batches: $\nabla Loss^{N}(\theta)= \sum_{b=1}^N \nabla Loss_b(\theta)$. For the training of 2D models we fixed N=10 and for the 3D case we fixed N=5 due to computational costs.

The second modification is to divide the gradient by its norm. With the update rule of the standard SGD, strong gradients would cause too high updates of the parameters which can even result in the divergence of the training and numerical problems. Conversely, weak gradients would result in too small updates and then a very slow training. We want therefore to be independent of the magnitude of the gradient in order to guarantee a stable training. To summarize, our update vector $v$ is computed as following: 
\begin{equation}
 v^{t+1}=\mu v^{t} - \alpha_t   \frac{\nabla Loss^{N} (\theta^{t})}{\| \nabla Loss^{N} (\theta^{t}) \|} 
\end{equation}

\begin{figure}[!ht]
\centering
\includegraphics[width=1.0\textwidth]{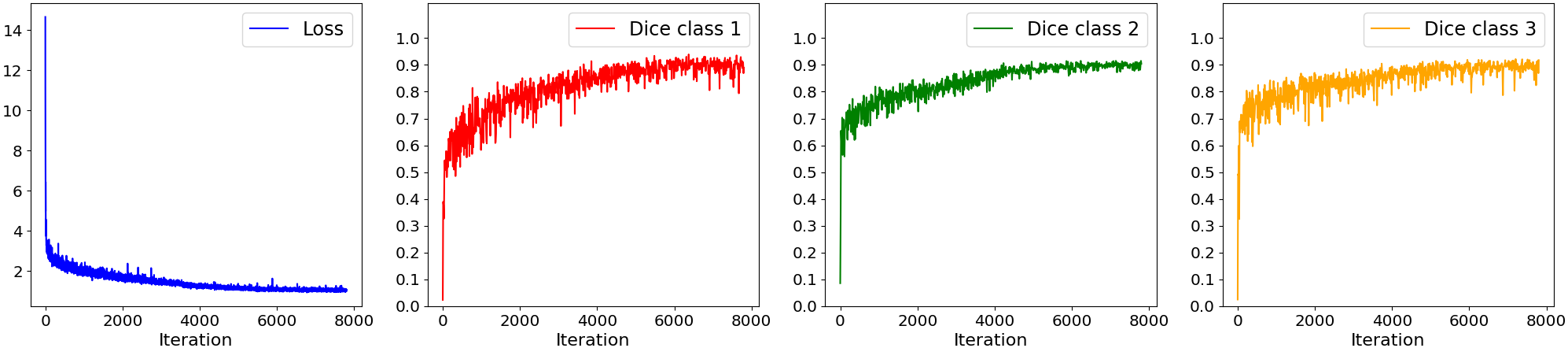}
\caption{Evolution of the loss and of Dice scores of tumor subclasses during the training of the 2D model.}
\label{fig_loss}
\end{figure}

In order to converge to a local minimum, we decrease the learning rate automatically according to the observed convergence speed. We fix the initial value $\alpha_{init}$ and the minimal value $\alpha_{min}$ of the learning rate. After each $F$ iterations we compute the mean loss accross the last $F/2$ iterations ($Loss_{current}$) and we compare it with the mean loss accross the previous $F/2$ iterations ($Loss_{previous}$) . We fix a threshold $0 < d_{loss} <1 $ on the relative decrease of the loss: if we observe $Loss_{current}> d_{loss} \times Loss_{previous} $ then the learning rate is updated as follows: $\alpha_{t+1}= max(\frac{\alpha_t}{2},\alpha_{min})$. Given that the loss is expected to decrease slower with the progress of the training, the value of F is doubled when we observe an insufficient decrease of the loss two times in a row. During the training of our models we fixed $\alpha_{init}=0.25$,  $\alpha_{min}=0.001$, $F=200$ and $d_{loss}=0.98$. The high values of the learning rate are due to the fact of dividing the gradients by their norm. The values of these hyperparameters were chosen empirically according to performed experiments. Fig. \ref{fig_loss} shows the evolution of the training loss of a 2D model along with Dice scores of tumor subclasses. This training scheme guaranteed a satisfactory convergence of our trainings while being easy to implement.

\section{Results}
\label{section_results}
\subsection{Data}

Our method was evaluated on datasets from the Multimodal Brain Tumor Segmentation Challenge (BRATS) 2017.  These datasets contain multisequence MR preoperative scans of patients diagnosed with malignant brain tumors. For each patient, four MR sequences were acquired: T1-weighted, post-contrast (gadolinium) T1-weighted, T2-weighted and FLAIR (Fluid Attenuated Inversion Recovery). The images come from 19 imaging centers and were acquired with different MR systems and with different clinical protocols. The images are provided after the pre-processing performed by the organizers: skull-stripped, registered to the same anatomical template and interpolated to $1mm^3$ resolution.

The ranges of image intensities highly vary between the scans due to image acquisition differences. We perform therefore a simple intensity normalization: for each patient and each MR sequence separately, we compute the median value of non-zero voxels, we divide the sequence by this median and we multiply it by a fixed constant. In fact, median is likely to be more stable than the mean, which can be easily impacted by the tumor zone. Experimentation with other normalization approaches such as histogram-matching methods \cite{nyul2000new} will be a part of the future work. Another potentially useful pre-processing could be bias field correction \cite{sled1998nonparametric}.

The ground truth corresponds to voxelwise annotations with 4 possible classes: non-tumor (class 0), contrast-enhancing tumor (class 3), necrotic and non-enhancing tumor (class 1), tumor-induced edema (class 2).

The \textit{Training} dataset contains 285 scans (215 high grade gliomas and 75 low grade gliomas) with provided ground truth segmentation. 
The \textit{Validation} dataset consists of 46 patients without provided segmentation and without provided information on the tumor grade. The evaluation on this dataset is performed via a public benchmark.

\subsection{Test setting}

We perform a series of experiments on two datasets of BRATS 2017 in order to analyze the effects of the main components of our method.

\begin{figure*}[!h]
\centering
\includegraphics[width=1.0\textwidth]{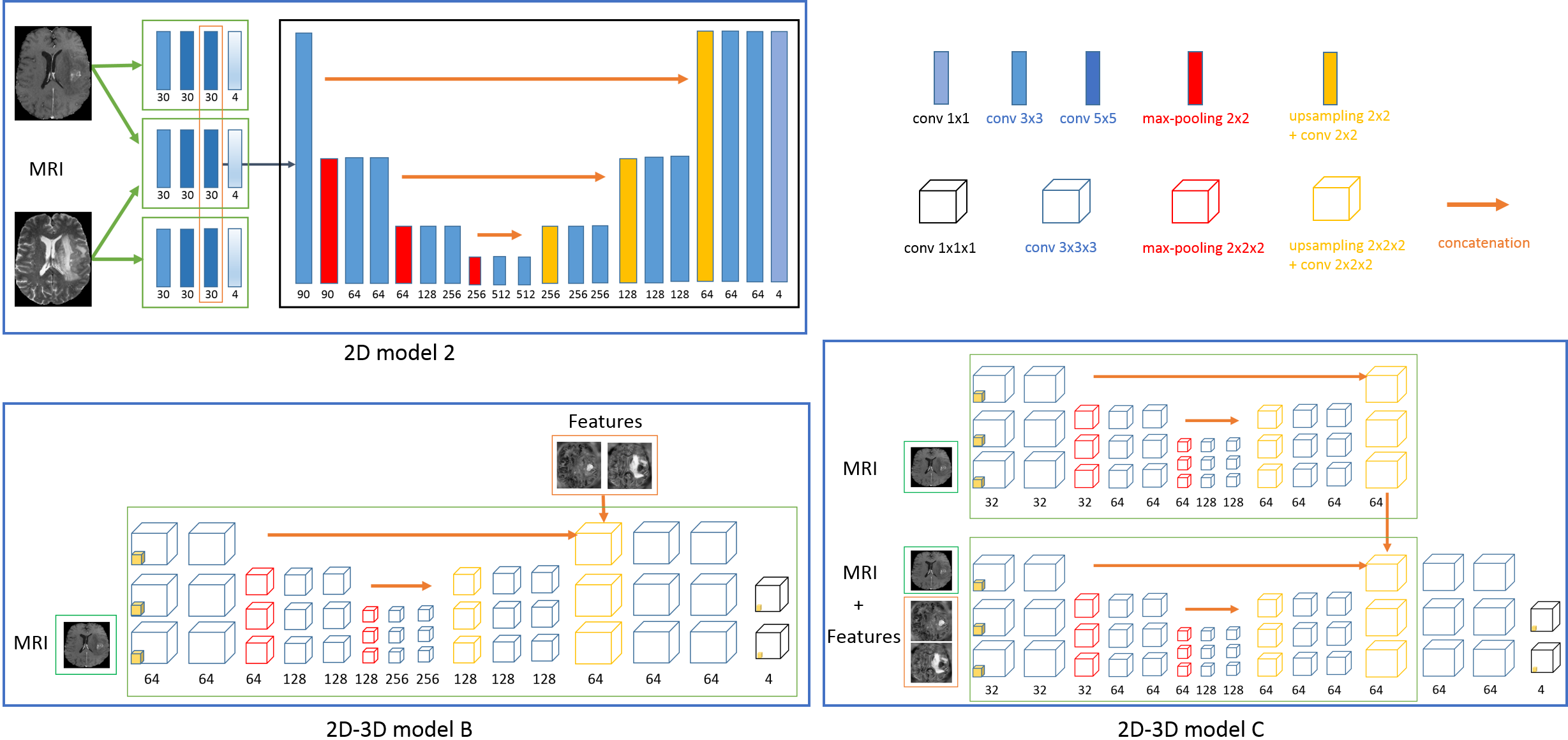}
\caption{Architectures of complementary networks used in our experiments.}
\label{fig_variants}
\end{figure*}

In our experiments we use two architectures of our 2D model and three architectures of the 2D-3D model. The main difference between the two 2D neural networks used in experiments is the architecture of subnetworks processing the input MR sequences. In the first 2D model, the subnetworks correspond to reduced versions of U-Net (Fig.~\ref{fig_model}) whereas in the second model, the subnetworks are composed of three convolutional layers (Fig. \ref{fig_variants}, top). In the remainder, we refer to these models as '2D model 1' and '2D model 2'. The difference between the two first 2D-3D models is the choice of the layer in which the 2D features are imported: in the first layer of the network (Fig. \ref{fig_model3D}) or before the final sequence of convolutional layers (Fig. \ref{fig_variants}, bottom left). The third 2D-3D model (Fig. \ref{fig_variants}, bottom right) is composed of two streams, one processing only the 3D image patch and the other stream taking also the 2D features as input. We refer to these models as 2D-3D model A, 2D-3D model B and 2D-3D model C. Please note that the two first models correspond to a standard 3D model with the only difference of taking an additional input.

Each of the 2D-3D models is trained twice using respectively the features learned by 2D model 1 or the features learned by 2D model 2. At the end we combine the 6 trained 2D-3D models with the voting strategy described in section \ref{section_ensembling}.

In the first series of experiments we qualitatively analyze the effects of using features learned by 2D networks as an additional input to 3D networks. The first test dataset is composed of 50 randomly chosen patients from the \textit{Training} dataset and the networks are trained on the remaining 235 patients. We refer to this dataset as 'local dataset' in the remainer. In the first step, 2D model 1 is trained separately on axial, coronal and sagittal slices and the standard 3D model is trained on 70x70x70 patches. Then we extract the features produced by the 2D model for all images of the training database and we train the same 3D model on 70x70x70 patches using these extracted features as an additional input (2D-3D model A specified on Fig. \ref{fig_model3D}).

Additionally, we test our architecture with modality-specific subnetworks in the context of missing MR sequences in the training database. In this setting, we suppose that the four MR sequences are available only for 20\% of patients and that for the remaining patients, one MR sequence out of the four is missing. More precisely, we randomly split the training set of 235 patients in five equal subsets (47 patients in each) and we consider that only the first subset contains all the four MR sequences whereas the four other subsets exclusively miss one MR sequence (T1, T1c, T2 or T2-FLAIR). We previously noted that modality-specific subnetworks can be trained independently: in this case, a subnetwork specific to a given MR sequence can be trained on 80\% of the training database (on all training images except the ones for which the MR sequence is missing). The goal of the experiment is to test if the training of these subnetworks improves the segmentation performance in practice. We first evaluate the performance obtained by 2D model 1 (version CNN-2DAxl) trained only on the training subset containing all MR sequences (47 patients). Then we evaluate the performance obtained when the subnetworks are pretrained, each of them using 80\% of the training database.

The second series of experiments is performed on \textit{Validation} set of BRATS 2017 composed of 46 patients without provided ground truth segmentation. We have evaluated our segmentation performance on the public benchmark of the challenge to compare our results with few dozens of teams from renowned research institutions worldwide.

In addition, on the two datasets we present the scores obtained by U-net processing axial slices, using our implementation (with batch-normalization).

The performance is measured by the Dice score between the segmentation $\tilde{Y}$ produced by the algorithm and the ground truth segmentation $Y$:
\begin{equation}
 DSC(\tilde{Y}, Y)= \frac{2 |\tilde{Y} \cap Y|}{|\tilde{Y}|+|Y|}
\end{equation}

We perform t-tests (paired, one-tailed) to measure statistical significance of the observed improvements provided by our three main contributions (2D-3D model, modality-specific subnetworks, merging strategy). We consider the significance level of 5\%.

\subsection{Results}

\begin{table}[ht!]
\centering
\caption{Mean Dice scores on the local dataset (50 patients).}
\begin{tabular}{|c|c|c|c|}
  \hline 
& \scriptsize{Enhancing Core}& \scriptsize{Tumor Core}& \scriptsize{Whole Tumor}\\ 
\hline
\scriptsize{Unet axial slices}& \scriptsize{ 73.9   }& \scriptsize{78.1}& \scriptsize{ 86.5  }\\ 
  \hline
\scriptsize{2D model 1 axial slices}& \scriptsize{  73.6 }& \scriptsize{ 79.4}& \scriptsize{ 86.6  }\\ 
  \hline
\scriptsize{Standard 3D model (without 2D features)}& \scriptsize{ 73.7 }& \scriptsize{ 77.0  }& \scriptsize{ 85.7  }\\ 
  \hline
\scriptsize{2D-3D model A, features from 2D model 1}& \scriptsize{ \textbf{77.4} }& \scriptsize{ \textbf{80.9} }& \scriptsize{ \textbf{87.3}   }\\ 
  \hline
\end{tabular}
\label{table_results}
\end{table}

\begin{table}[ht!]
\centering
\caption{Mean Dice scores on the local dataset (50 patients) with misssing MR sequences. EC, TC and WT refer respectively to 'Enhancing Core', 'Tumor Core' and 'Whole Tumor' regions.}
\begin{tabular}{|c|c|c|c|}
  \hline
& \scriptsize{EC }& \scriptsize{TC}& \scriptsize{WT}\\  
\hline
\scriptsize{  2D model 1, missing data}& \scriptsize{  70.2 }& \scriptsize{  68.6}& \scriptsize{ 83.0  }\\ 
  \hline
\scriptsize{ 2D model 1 missing data + pretrained subnetworks}& \scriptsize{ \textbf{71.9} }& \scriptsize{ \textbf{73.7}  }& \scriptsize{ \textbf{84.1}  }\\ 
    \hline
\scriptsize{  2D model 1 full data}& \scriptsize{  73.6 }& \scriptsize{ 79.4}& \scriptsize{ 86.6  }\\ 
  \hline
\end{tabular}
\label{table_results_subnetworks}
\end{table}

\begin{table}[ht!]
\centering
\caption{Mean Dice scores on the Validation set of BRATS 2017 (46 patients).}
\begin{tabular}{|c|c|c|c|}
  \hline
& \scriptsize{EC}& \scriptsize{TC}& \scriptsize{WT}\\ 
\hline
\scriptsize{Unet axial slices}& \scriptsize{ 71.4   }& \scriptsize{ 76.6}& \scriptsize{ 87.7  }\\ 
  \hline
\scriptsize{2D model 1 axial slices}& \scriptsize{  71.1 }& \scriptsize{ 78.4}& \scriptsize{ 88.6  }\\ 
\hline
\scriptsize{2D model 2 axial slices}& \scriptsize{ 68.0 }& \scriptsize{ 78.3}& \scriptsize{ 88.1   }\\ 
  \hline
\scriptsize{Standard 3D model (without 2D features)}& \scriptsize{ 68.7 }& \scriptsize{ 74.2  }& \scriptsize{ 85.4  }\\ 
  \hline
\scriptsize{* 2D-3D model A, features from 2D model 1}& \scriptsize{ 76.7 }& \scriptsize{ 79.5 }& \scriptsize{ 89.3   }\\ 
  \hline
\scriptsize{* 2D-3D model B, features from 2D model 1}& \scriptsize{ 76.6 }& \scriptsize{ 79.1 }& \scriptsize{ 89.1  }\\ 
  \hline
\scriptsize{* 2D-3D model C, features from 2D model 1}& \scriptsize{ 76.9 }& \scriptsize{ 78.3 }& \scriptsize{ 89.4   }\\ 
  \hline
\scriptsize{*  2D-3D model A, features from 2D model 2}& \scriptsize{ 73.4 }& \scriptsize{ 79.5 }& \scriptsize{ 89.7  }\\ 
 \hline
\scriptsize{*  2D-3D model B, features from 2D model 2}& \scriptsize{ 74.1 }& \scriptsize{ 79.4 }& \scriptsize{ 89.5   }\\ 
  \hline
\scriptsize{* 2D-3D model C, features from 2D model 2}& \scriptsize{ 74.3 }& \scriptsize{ 79.4 }& \scriptsize{ 89.6 }\\ 
  \hline
\scriptsize{ Final segmentation: ensembling of models *}& \scriptsize{ \textbf{77.2} }& \scriptsize{ \textbf{80.8} }& \scriptsize{ \textbf{90.0}  }\\ 
  \hline

\end{tabular}
\label{table_results2}
\end{table}

\begin{table}[ht!]
\centering
\caption{p-values of the t-tests (in bold: statistically significant results) of the improvement provided by our main contributions. To lighten the notations, '2D' refers to '2D model 1 axial slices' and '2D-3D' refers to '2D-3D model A, features from 2D model 1'.}
\begin{tabular}{|c|c|c|c|}
\hline

  \hline

&   \scriptsize{EC} &  \scriptsize{TC} &  \scriptsize{WT}    \\ 
  \hline

\scriptsize{2D vs 2D with pretrained subnetworks, missing data}   &  \scriptsize{\textbf{0.0054}} &  \scriptsize{\textbf{0.0003}}&  \scriptsize{\textbf{0.0074}}  \\

  \hline

  \scriptsize{Standard 3D vs 2D-3D, dataset 1}  &  \scriptsize{\textbf{0.0082}}&  \scriptsize{\textbf{0.0016}}&  \scriptsize{0.0729}  \\

\hline

\scriptsize{Standard 3D vs 2D-3D, dataset 2} &  \scriptsize{\textbf{0.0077}}&  \scriptsize{\textbf{0.0005}}&  $\scriptsize{\textbf{$<$0.0001}}$  

\\

  \hline

\scriptsize{2D-3D vs ensembling of 2D-3D}   &  \scriptsize{0.1058}&  \scriptsize{\textbf{0.0138}}&  \scriptsize{\textbf{0.0496}}  \\ 
  \hline

\end{tabular}
\label{table_pvalues}
\end{table}

\begin{figure}[ht!]
\centering
\includegraphics[width=0.98\textwidth]{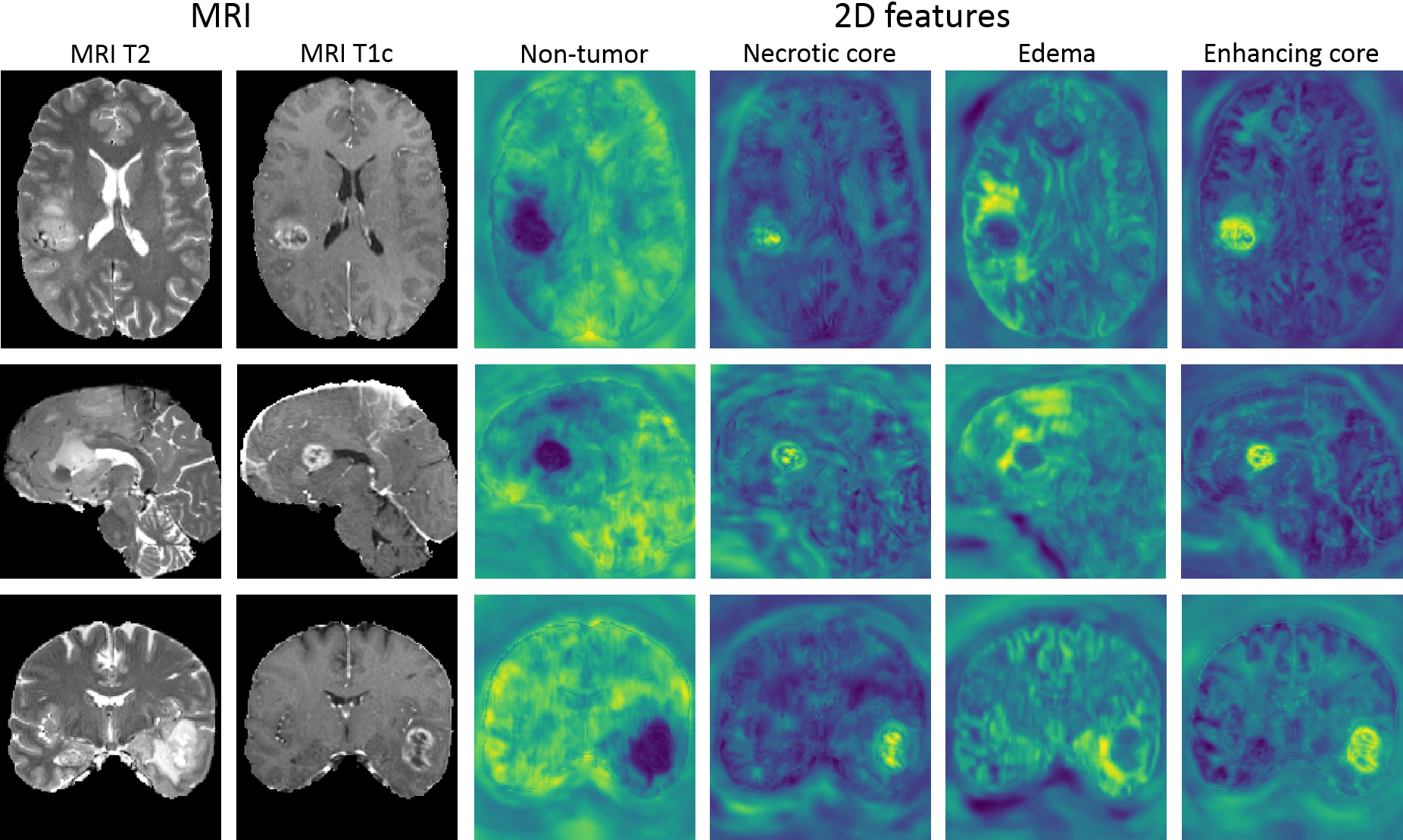}
\caption{2D features computed for three different patients from the test set. These features correspond to unnormalized outputs of the final convolutional layers of three versions of a 2D model (CNN-2DAxl, CNN-2DSag, CNN-2DCor). The values of these features are used as an additional input to a 3D CNN. Each feature highlights one of the tumor classes (columns 3-6) and encodes a rich information extracted from a long-range 2D context within an axial, sagittal or coronal plane (rows 1-3). Each row displays a different case from the test set (unseen by the network during the training).}
\label{fig_features}
\end{figure}

\begin{table}[ht!]
\centering
\caption{Mean Dice scores of the 10 best scoring teams on the validation leaderboard of BRATS 2017 (state of January 22, 2018)}
\begin{tabular}{|c|c|c|c|c|c|c|c|}

  \hline
& \scriptsize{EC}& \scriptsize{TC}& \scriptsize{WT}& \scriptsize{Rank EC}& \scriptsize{Rank TC}& \scriptsize{Rank WT}& \scriptsize{Average rank}\\ 
\hline
\scriptsize{UCL-TIG }& \scriptsize{ 78.6 }& \scriptsize{ 83.8 }& \scriptsize{ 90.5 }& \scriptsize{ 1   / 55 }& \scriptsize{ 1 }& \scriptsize{ 1 }& \scriptsize{ 1.0 }\\ 
 \hline 
\scriptsize{MIC\_DKFZ }& \scriptsize{ 77.6 }& \scriptsize{ 81.9 }& \scriptsize{ 90.3 }& \scriptsize{ 2   / 55 }& \scriptsize{ 2 }& \scriptsize{ 2 }& \scriptsize{ 2.0 }\\ 
 \hline 
\scriptsize{\textbf{inpm (our method)} }& \scriptsize{ 77.2 }& \scriptsize{ 80.8 }& \scriptsize{ 90.0 }& \scriptsize{ 3   / 55 }& \scriptsize{ 3 }& \scriptsize{ 7 }& \scriptsize{ 4.3 }\\ 
 \hline 
\scriptsize{UCLM\_UBERN }& \scriptsize{ 74.9 }& \scriptsize{ 79.1 }& \scriptsize{ 90.1 }& \scriptsize{ 9   / 55 }& \scriptsize{ 6 }& \scriptsize{ 3 }& \scriptsize{ 6.0 }\\ 
\hline 
\scriptsize{biomedia1 }& \scriptsize{ 73.8 }& \scriptsize{ 79.7 }& \scriptsize{ 90.1 }& \scriptsize{ 12 / 55 }& \scriptsize{ 5 }& \scriptsize{ 5 }& \scriptsize{ 7.3 }\\ 
 \hline 
\scriptsize{stryker }& \scriptsize{ 75.5 }& \scriptsize{ 78.3 }& \scriptsize{ 90.1 }& \scriptsize{ 6   / 55 }& \scriptsize{ 10 }& \scriptsize{ 6 }& \scriptsize{ 7.3 }\\ 
 \hline 
\scriptsize{xfeng }& \scriptsize{ 75.1 }& \scriptsize{ 79.9 }& \scriptsize{ 89.2 }& \scriptsize{ 8   / 55 }& \scriptsize{ 4 }& \scriptsize{ 11 }& \scriptsize{ 7.7 }\\ 
 \hline 
\scriptsize{Zhouch }& \scriptsize{ 75.4 }& \scriptsize{ 77.8 }& \scriptsize{ 90.1 }& \scriptsize{ 7   / 55 }& \scriptsize{ 12 }& \scriptsize{ 4 }& \scriptsize{ 7.7 }\\ 
 \hline 
\scriptsize{tkuan }& \scriptsize{ 76.5 }& \scriptsize{ 78.2 }& \scriptsize{ 88.9 }& \scriptsize{ 4   / 55 }& \scriptsize{ 11 }& \scriptsize{ 13 }& \scriptsize{ 9.3 }\\ 
 \hline 
\scriptsize{Zhao }& \scriptsize{ 75.9 }& \scriptsize{ 78.9 }& \scriptsize{ 87.2 }& \scriptsize{ 5  / 55 }& \scriptsize{ 7 }& \scriptsize{ 16 }& \scriptsize{ 9.3 }\\ 
 \hline 
\end{tabular}
\label{table_brats}
\end{table}

\begin{table}[ht!]
\centering
\caption{Distribution of Dice scores (final result). The numbers in brackets denote standard deviations.}
\begin{tabular}{|c|c|c|c|}
  \hline
& \scriptsize{Enhancing Core}& \scriptsize{Tumor Core}& \scriptsize{Whole Tumor}\\ 
  \hline
\scriptsize{Mean}& \scriptsize{ 77.2 (24.4)   }& \scriptsize{ 80.8 (18.9) }& \scriptsize{   90.0 (8.1)  }\\ 
\hline
\scriptsize{Median }& \scriptsize{ 85.4   }& \scriptsize{ 88.3}& \scriptsize{  91.8   }\\ 
  \hline
\scriptsize{Quantile 25 \%   }& \scriptsize{ 76.9 }& \scriptsize{  75.0 }& \scriptsize{    89.6 }\\ 
  \hline
\scriptsize{Quantile 75 \%      }& \scriptsize{ 90.0 }& \scriptsize{ 93.5 }& \scriptsize{   94.5  }\\ 
  \hline

\end{tabular}
\label{table_stats}
\end{table}

\begin{figure*}[ht!]
\centering
\includegraphics[width=0.98\textwidth]{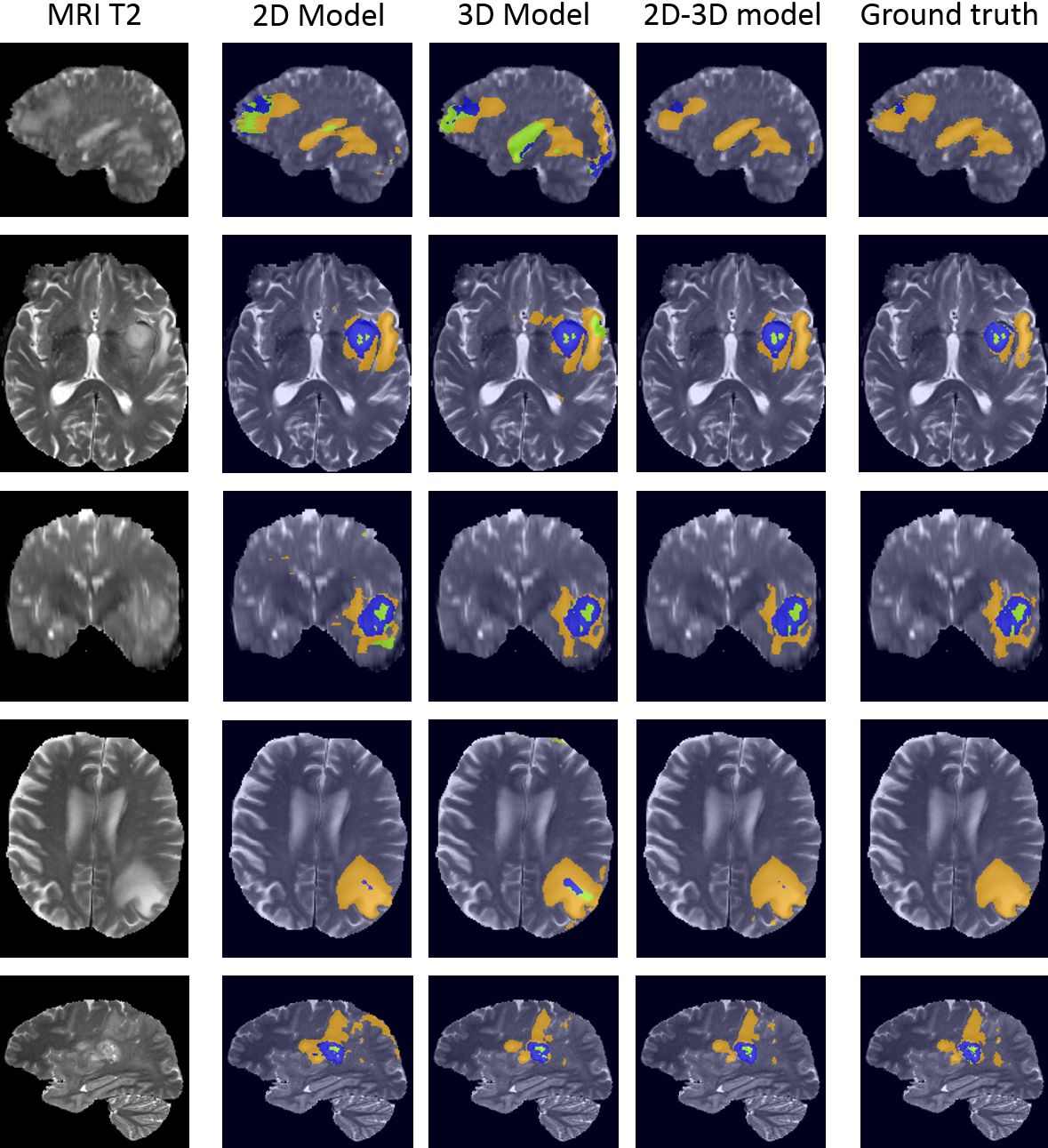}
\caption{
Examples of segmentations obtained with models using a different spatial context. Each row represents a different patient from the local test dataset (images unseen during the training). From left to right: MRI T2, '2D model 1' processing the image by axial slices, standard 3D model (without 2D features), '2D-3D model A' using the features produced by '2D model 1', ground truth segmentation.}
\label{fig_results}
\end{figure*}

First, we observe that the use of features produced by 2D neural networks improves the performance of 3D models. In all experiments the 3D models using features extracted by 2D networks obtained better performance than their standard 3D counterparts and than the 2D neural networks from which the features were extracted (Table \ref{table_results} and Table \ref{table_results2}). 

The second important observation are the high Dice scores obtained by our main 2D model processing the image by axial slices without any postprocessing. This result is particularly encouraging given that our 2D system is easier to use in real clinical settings than 3D models (which implicitly assume a fixed resolution in three dimensions for all patients) and can be trained on larger databases than standard models, being more robust to the problem of missing MR sequences due to its architecture with modality-specific subnetworks.

The advantages of our architecture with modality-specific subnetworks are demonstrated by the test performed in the setting with missing MR sequences (Table \ref{table_results_subnetworks}). In this case, each modality-specific subnetwork can be naturally trained on a large part of the training database: on all training cases except the ones for which its MR sequence is missing. Pretraining of these modality-specific subnetworks improved the segmentation performance on the test set for all tumor subregions (Table \ref{table_results_subnetworks}). Even if the multiclass segmentation problem is very difficult for a small network using only one MR sequence, this pretraining forces the subnetwork to learn the most relevant features, which will then be used by the main part of the network, trained on the subset of training cases for which all MR sequences are available. The improvement was found statistically significant for all the three tumor subregions (Table \ref{table_pvalues}).

The qualitative analysis (Fig. \ref{fig_results}) of outputs of 2D neural networks highlights two main problems of 2D approches. First, as expected, the produced segmentations show discontinuities which appear as patterns parallel to the planes of processing. The second problem are false positives in the slices at the borders of the brain and containing artefacts of skull-stripping.

The segmentations produced by the standard 3D model are more spatially consistent but the network suffers from a limited input information from distant voxels. The use of learned features (displayed on Fig. \ref{fig_features} for 2D model 1 trained on the first dataset) as an additional input to the network gives a considerable advantage by providing rich information extracted from distant points. The difference of performance is particulary visible for 'tumor core' and 'enhancing core' subregions. 

The ensembling of models with our decision rule further improves the segmentation accuracy for all tumor subregions. This improvement was found statistically significant for 'whole tumor' and 'tumor core' subregions.

Our method compares favorably with competing methods of BRATS 2017 (Table \ref{table_brats}): among 55 teams which evaluated their methods on all test patients of the validation set, we obtain top-3 performance for 'core' and 'enhancing core' tumor subregions. We obtain mean Dice score of 0.9 for the 'whole tumor' region, which is almost equal to the one obtained by the best scoring team (0.905). The leaderboard of BRATS 2017 only shows mean performances obtained by participating teams. However, the benchmark individually provides detailed scores and complementary statistics, in particular quartiles and standard deviations reported in Table \ref{table_stats}. Our method yields promising results with median Dice score of 0.918 for the \textit{whole tumor}, 0.883 for the \textit{tumor core} and 0.854 for the \textit{enhancing core}. While the Dice scores for the \textit{whole tumor} region are rather stable (generally between 0.89 and 0.95), we observe a high variability of the scores obtained for the tumor subregions. In particular the obtained median Dices are much higher than the means, due to the sensitivity of Dice score to outliers.

\section{Discussion and conclusion}
In this paper we presented a deep learning system for multiclass segmentation of tumors in large multisequence MR scans. The goal of our work was to propose elements to improve the performance, the robustness and the applicability of commonly used CNN-based systems. In particular, we proposed a new methodology to capture a long-range 3D context with CNNs, we introduced a network architecture with modality-specific subnetworks and we proposed a new voting strategy to merge multiclass segmentations produced by different models.

First, we proposed to use features learned by 2D CNNs (capturing a long-range 2D context in three orthogonal directions) as an additional input to a 3D CNN. Our approach combines the strengths of 2D and 3D CNNs and was designed to capture a very large spatial context while being efficient in terms of computations and memory load. Our experiments showed that this hybrid 2D-3D model obtains better performances than both the standard 3D approach (considering only the intensities of voxels of a subvolume) and than the 2D models which produced the features. Even if the use of the additional input implies supplementary reading operations, the simple importation of few features to a CNN does not considerably increase the number of computations and the memory load. In fact, in typical CNNs performing hundreds of convolutions, max-poolings and upsamplings, the data layer represents typically a very small part of the memory load of the network. One solution to limit the reading operations could be to read downsampled versions of features or to design a 2D-3D architecture in which the features are imported in a part of the network where the feature maps are relatively small.

The improvement provided by the 2D-3D approach has the cost of increasing the complexity of the method compared to a pure 3D approach as it requires a two-step processing (first 2D, then 3D). However, its implementation is rather simple as the only supplementary element to implement is the extraction of 2D features, i.e. computation of outputs of trained 2D networks (with a deep learning software such as TensorFlow) and saving the obtained tensors in files. In the 3D part, the extracted features are then simply read as additional channels of the input image.

Despite the important recent progress of GPUs, pure 3D approaches may be easily limited by their computational requirements when the segmentation problem involves an analysis of a very large spatial 3D context. In fact, Convolutional Neural Networks require an important amount of GPU memory and a high computational power as they perform thousands of costly operations on images (convolutions, max-poolings, upsamplings). The main advantage of our 2D-3D approach is to considerably increase the size of the receptive field of the model while being efficient in terms of the computational load. The use of our 2D-3D model may therefore be particularly relevant in the case of very large 3D scans.

Second, we proposed a novel approach to process different MR sequences, using an architecture with modality-specific subnetworks. Such design has the considerable advantage of offering a possibility to train one part of the network on databases containing images with missing MR sequences. 
In our experiments, training of modality-specific subnetworks improved the segmentation performance in the setting with missing MR sequences in the training database. Moreover, the fact that our 2D model obtained promising segmentation performance is particularly encouraging given that 2D networks are easier to apply for the clinical use where images have a variable number of acquired slices. Our approach can be easily used with any deep learning software (e.g. Keras). In the case of databases with missing MR sequences, the user only has to perform a training of a subnetwork (on images for which the given MR sequence is provided) and then read the learned parameters for the training of the main part of the network (on images for which all MR sequences are available).

In order to be less prone to limitations of particular choices of neural network architectures, we proposed to merge outputs of several models by a voxelwise voting strategy taking into account the semantics of labels. Our merging strategy can be naturally applied to any multiclass segmentation problem and allows the user to control the sensitivity to different tumor subregions. This possibility is important as false negatives (non-detections) of lesions are particularly problematic in the clinical setting. However, the use of a merging strategy comes with an additional complexity of the method and the user may prefer to use an unique network architecture.

In constrast to most methods, we do not apply any postprocessing on the produced segmentations.

Our three main contributions lead to introduction of new hyperparameters (weights of the loss function, thresholds for merging of models). These hyperparameters are normalized (numbers between 0 and 1) and their tuning corresponds to controlling the importance of different elements, in particular the sensitivity to different tumor subclasses.

Our methodological contributions can be easily included separately or jointly into a CNN-based system to solve specific segmentation problems.

An interesting direction for the future work would be weakly-supervised or semi-supervised learning. In fact, an accurate manual segmentation of tumors is difficult to obtain due to the labour and competences necessary to perform it. Approaches which rely less on manual segmentations are therefore of particular interest.

\section*{Acknowledgements}
Pawel Mlynarski is funded be Microsoft Research-INRIA Joint Centre, France. This work was supported by the Inria Sophia Antipolis - M\'editerran\'ee, "NEF" computation cluster.









\bibliographystyle{elsarticle-num-names} 
\bibliography{biblio}

\end{document}